\documentclass[preprint, authoryear, 3p, a4paper, lefttitle, twocolumn]{elsarticle}
\usepackage[utf8]{inputenc}
\usepackage[T1]{fontenc}
\usepackage{placeins}
\usepackage{amsmath}
\usepackage{amsfonts}
\usepackage{amssymb}
\usepackage{graphicx}
\newsavebox\CBox 
\usepackage{booktabs}
\usepackage{hyperref}
\usepackage{cleveref}
\usepackage{subcaption}
\usepackage{tabularx}
\usepackage{printlen}
\usepackage{stfloats}
\usepackage{enumitem}
\usepackage{adjustbox}
\usepackage{xcolor}

\usepackage{etoolbox}

\newcolumntype{L}{>{\raggedright\arraybackslash}X}
\newcolumntype{C}{>{\centering\arraybackslash}X}
\newcolumntype{R}{>{\raggedleft\arraybackslash}X}

\newcommand*\fr[1]{\sbox\CBox{#1}\resizebox{\wd\CBox}{\ht\CBox}{\textbf{#1}}}
\newcommand*\sr[1]{\underline{#1}}

\title{Facilitated machine learning for image-based fruit quality assessment}

\author[1,2,3]{Manuel Knott}
\ead{manuel.knott@empa.ch}

\author[2,3]{Fernando Perez-Cruz}
\ead{fernando.perezcruz@sdsc.ethz.ch}

\author[1]{Thijs Defraeye\corref{cor1}}
\ead{thijs.defraeye@empa.ch}

\cortext[cor1]{Corresponding author: thijs.defraeye@empa.ch}
\address[1]{Empa, Swiss Federal Laboratories for
Materials Science and Technology, Laboratory for Biomimetic Membranes and Textiles, St. Gallen, Switzerland}
\address[2]{Swiss Data Science Center, ETH Zurich and EPFL, Zurich, Switzerland}
\address[3]{Institute for Machine Learning, Department of Computer Science, ETH Zurich, Switzerland}

\begin{document}

\emergencystretch 3em
\begin{abstract}
Image-based machine learning models can be used to make the sorting and grading of agricultural products more efficient. In many regions, implementing such systems can be difficult due to the lack of centralization and automation of postharvest supply chains. Stakeholders are often too small to specialize in machine learning, and large training data sets are unavailable.
We propose a machine learning procedure for images based on pre-trained Vision Transformers. It is easier to implement than the current standard approach of training Convolutional Neural Networks (CNNs) as we do not (re-)train deep neural networks.
We evaluate our approach based on two data sets for apple defect detection and banana ripeness estimation. Our model achieves a competitive classification accuracy equal to or less than one percent below the best-performing CNN. At the same time, it requires three times fewer training samples to achieve a 90\% accuracy.
\end{abstract}

\maketitle
\section{Introduction}
The automated quality assessment of fruit images by computer vision is commonplace in the postharvest supply chain and is typically used for sorting and grading. Sorting is a selection based on a single attribute, such as size or color. Grading is the classification according to several quality attributes such as color, size, or defects \citep{abbas_automated_2019}. They provide several advantages for packhouse operators and retailers in the food supply chain. First, damaged or off-spec fresh produce is sorted out. Fruits can also be packed better if they are sorted based on size, for example, in trays or boxes.  
In addition, grading of fruits and vegetables into different categories based on quality at harvest enables producers to sell their produce in different price ranges for the different grades based on consumers' needs.
In case of limited cooling capacity, producers can also choose to store only high-value products in cold storage. Such automated sorting and grading largely replaced manual quality estimation for fresh produce in high-income countries due to its increased efficiency and reduced labor cost. At the same time, algorithm-based solutions establish a common standard for visual inspection and remove human bias \citep{abbas_automated_2019}.

Color spectrometers are a direct and precise way to assess color-based fruit properties such as its ripeness level \citep{alander_review_2013, das_ultra-portable_2016}. However, assessing more complex visual features -- such as surface patterns that correlate with specific defects -- often requires the application of Computer Vision based on RGB images taken with regular cameras.
Early approaches rely on image processing and manual feature crafting often combined with shallow machine learning models \citep[e.g.][]{dubey_detection_2012, prabha_assessment_2015, piedad_postharvest_2018, mazen_ripeness_2019, santos_pereira_predicting_2018}. \citet{khan_ccdf_2018} combine manual features with deep features from a pre-trained VGG neural network and classify them using a Support Vector Machine. Later approaches directly train neural networks on smaller domain data sets \citep[e.g.][]{selvaraj_ai-powered_2019, lopes_deep_2022}.
\citet{ismail_real-time_2022} achieve excellent classification results for the data sets that are also used in this research by using sophisticated image pre-processing and neural networks.
Convolutional Neural Networks (CNNs) are the established approach for machine learning on image data and are used in numerous applications in different fields. Training such a model from scratch is not feasible for most use cases, as it requires a large number of labeled training samples, enhanced computational resources, and extensive experimentation to find the best model setup. Therefore, machine learning practitioners commonly make use of transfer learning: Instead of starting from scratch, one uses a model that is already pre-trained on a large general data set and finetunes it on a smaller domain-specific data set \citep{ribani_survey_2019, saranya_banana_2021}. While this procedure reduces the required amount of training data drastically, enhanced computational resources in form of GPUs are still required. Furthermore, the required number of labeled training samples can still be high, especially when the task to learn is difficult. For postharvest sorting and grading applications, retraining can be required for some applications to account for fruit quality differences every harvest season. In summary, CNN-based machine learning for automated sorting and grading requires large labeled image data sets and extensive computational resources in the form of GPUs for (re-)training models.

While large centralized fruit sorting lines readily meet these requirements, this does often not apply to regions with decentralized food supply chains where sorting and grading are done manually \citep{abbas_automated_2019}.

Supply chains in remote areas or in developing countries can often be characterized by stakeholders that are too small to specialize in high-tech solutions.
According to \citet{de-arteaga_machine_2018} this leads to several challenges when implementing machine learning solutions. Three of those directly apply to computer vision for quality assessment in postharvest supply chains: 
\begin{enumerate}
    \item \textit{Learning with small data sets:} Due to the lack of highly automated sorting lines, images are often sourced by cold-chain operators, for example, with smartphone cameras in a non-standardized way. This naturally limits the sizes of data sets and the quality of the images.
    \item \textit{Learning from multiple messy data sets:} To cover different crops at different ripeness levels, images need to be taken at different times of the season or at different stages in the supply chain. This non-centralized and non-standardized data acquisition creates additional challenges. One must address the risk of introducing unwanted biases when combining data from different sources.
    \item \textit{Learning with limited computational resources:} Farmers, packhouse operators, and retailers are often too small and not specialized enough, so investment into advanced computational resources or cloud services would be worthwhile. By identifying machine learning solutions that can be developed on standard hardware, the barrier to using such technologies can be lowered.
\end{enumerate}

This research presents an approach focusing on the issues of ``small data sets'' and ``limited computational resources''.
We make use of Vision Transformers (ViTs) that are pre-trained in a self-supervised manner. 
We use those models as ``off-the-shelf'' feature generators that translate high-dimensional and unstructured image data into a lower-dimensional domain. Those feature representations (``embeddings'') can then be input to simpler machine learning models, which can be trained with minimal effort on low-cost hardware to perform the desired classification or regression task. In this procedure, there is no need for resource-intensive deep learning on the domain data set, which addresses the issue of limited computational resources. 
We demonstrate the competitiveness of this approach by benchmarking two different classification tasks for estimating banana ripeness and defective apples, respectively.
In addition, we show that our method is superior when only a small amount of labeled training data is available. Therefore, we address the need for learning algorithms that work on limited data.
%


\section{Materials and Methods}

\subsection{Data sets}
\label{sec:datasets}

This study involves data sets that address two types of visual fruit quality attributes. The first is color-based, which often relates to a fruit's ripeness. Here we target banana fruits. The second is based on local features, such as surface defects or surface spots. These traits indicate, for example, insect damage. Here we target apple fruits. As such, our study includes one representative data set for each of those challenges.

The banana ripeness data set from Fayoum University \citep{mazen_ripeness_2019} (short: ``Fayoum Banana'') contains 273 single fruit images of bananas of different ripeness levels with a neutral background. Even though the ripeness degree of a banana is a continuous scale, in reality, farmers and cold store managers grade the fruit using a scale from 1 to 7. The researchers who provided this data set labeled it into four ordinal classes: ``green'' (104 samples), ``yellowish-green'' (48 samples), ``midripen'' (88 samples), and ``overripen'' (33 samples). However, using ordinal classes naturally leads to edge cases where samples can hardly be objectively associated with one class. Consequently, this leads to a human-induced annotation error. Our own visual examination of the images disclosed possibly inaccurate labels, especially between the yellowish-green and midripen classes.
From a machine learning perspective, this classification problem can be considered an easy task since the classification is mainly dependent on the average color values.

For apple fruits, the Internal Feeding Worm data set of the Comprehensive Automation for Specialty Crops research project \citep{li_casc_2009} (short: ``CASC~IFW'') was used. It contains 5858 images of apples with and without worm defects at the apple surface. Images were taken from four apple cultivars on trees from June to September and, therefore, at different ripeness levels ranging from green to red apples. We use it for binary classification and distinguish between ``healthy'' (2058 samples) and ``damaged'' (3800 samples) fruits. The images have varying lighting conditions and backgrounds. The size and location of the feature of interest --- the damaged spots --- varies. It can be unremarkable and hard to differentiate from the apple's calyx. For those reasons, this classification problem can be considered an advanced machine learning task.

Since we are using pre-trained models, our data sets require preprocessing to match the resolution that was used for pre-training.
The CASC~IFW images were upsampled from originally 120x120 pixels to 224x224 using bilinear interpolation.
The Fayoum Banana images were downsampled to 224x224 pixels keeping the images' original ratio by applying zero-padding to the height dimension (top and bottom side).
Example images for both data sets can be found in \ref{sec:appendixA}.

\subsection{Machine Learning with Image Data}

Images are a high-dimensional and highly unstructured type of data representation.
Traditional (shallow) machine learning models fail to automatically capture the spatial relations of neighboring pixels in an image. Therefore, early approaches rely on extracting hand-crafted features (e.g., color histograms) to make shallow models work with images. However, these approaches are unsuitable for complex tasks and require a lot of manual work by domain experts.

In recent years, Deep Learning models, i.e., Convolutional Neural Networks (CNNs), have been established as the standard approach for image-based machine learning. These models can learn complex patterns in images in a self-reliant way. In simplified terms, a CNN for image classification consists of two high-level building blocks (see \Cref{fig:concept}, top): 
\begin{enumerate}
    \item A \textit{feature extractor} --- usually a combination of Convolutional and Pooling layers --- that translates an image into a lower-dimensional representation that is easier to work with for subsequent tasks. This part contains a large number of trainable parameters and requires large data sets to be trained on. In this study, we will use a pre-trained feature extractor.
    \item A \textit{classifier} that uses the latent features to solve the desired downstream task (e.g., finding the most probable affiliation in a given set of classes). It usually consists of one to only a few simple feed-forward layers and is, therefore, fast to train.
\end{enumerate}

It is common practice to combine pre-trained feature extractors with untrained classifiers for faster convergence on new machine learning tasks. This concept, called ``transfer learning'' accelerates the development of new models significantly.
However, the latent features can be hyper-optimized on the task and/or data set the model was pre-trained on. Therefore, to get satisfying results with a new smaller domain data set, one often needs to at least fine-tune the feature extractor's parameters. In that case, the required training time is still significantly lower compared to training a model from scratch, but advanced computational resources can still be required in the form of GPUs. As detailed below, we can also use pre-trained feature extractors without re-training or fine-tuning, which we target in this work. \citep{ribani_survey_2019}

\subsubsection{Vision Transformers}

\begin{figure*}[tb!]
    \centering
    \includegraphics[width=\textwidth]{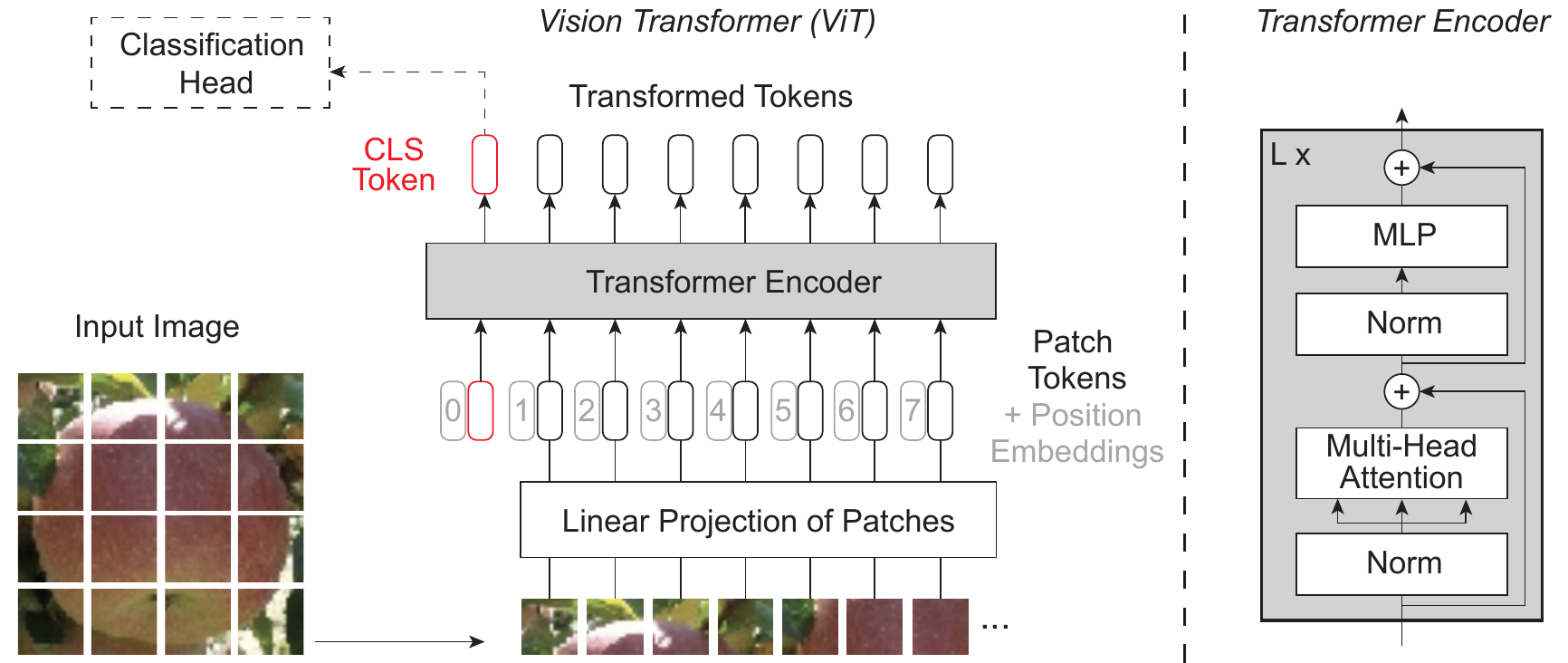}
    \caption{Model overview of a vision transformer. An input image is split into patches and embedded with positional information (``patch tokens''). An extra learnable \texttt{cls}-token is added. All tokens are transformed into a new sequence by a transformer encoder. In a classification setup, an additional simple sub-model (``classification head'') learns the class based on the transformed \texttt{cls}-token. The illustration is adapted from \citet{dosovitskiy_image_2020}.}
    \label{fig:vit}
\end{figure*}

In this research, we propose an alternative method for image classification with small data sets. It is built upon a pre-trained Vision Transformer.
The Transformer architecture as shown in \Cref{fig:vit} was originally introduced by \citet{vaswani_attention_2017} and has been established as the standard method for Natural Language Processing tasks, such as machine translation or text classification. Later variants use attention in a self-supervised way, which enables the use of large unlabelled text corpora as training data \citep{devlin_bert_2019, radford_language_2019}.
Recent advances in computer vision have successfully adapted these models for image data with only small adaptions to the original approach \citep{dosovitskiy_image_2020, touvron_training_2021}.
Unlike CNNs, transformers can model long-range dependencies and are not limited to local features. Also, they scale well to large network capacities and data sets. \citep{khan_transformers_2021}

In the Vision Transformer (ViT) architecture, an input image is split into patches, while each patch is encoded into an embedding representation using a trainable linear projection.
These patch embeddings are called ``tokens,'' and they are the equivalent of words in a natural language context.
The sequence of flattened one-dimensional tokens is extended by positional encoding to retain positional information. The key component of the architecture -- the Transformer Encoder -- acts as a sequence-to-sequence translator and, therefore, generates a new output sequence of identical length and shape. 
The standard approach is to extend the patch tokens by an additional learnable class (``cls'') token. In a supervised classification setup, this token comprises the learned class-related features. This condensed vector is translated to a class label by a simple Multilayer Perceptron\footnote{A feed-forward neural network with a small number of hidden layers} (``classification head''), equivalent to the classifier in a CNN.

The key component in the transformer encoder is the \textit{self-attention} mechanism. Attention in machine learning describes a method of dynamically weighting inputs based on their actual value instead of weighting them equally. This is realized by adapting the key-query-value concept from retrieval systems. All three components are implemented as vectors and mapped to an output which is the weighted sum of the values. Self-attention in transformers is usually a ``scaled dot-product attention'' as originally introduced by \citet{vaswani_attention_2017}. It can be calculated as follows, while $d_k$ describes the number of key dimensions:

\begin{equation}
\mathrm{Attention}(Q, K, V) = \mathrm{softmax}\left( \dfrac{QK^\top}{\sqrt{d_k}}\right)V
\end{equation}

In practice, it has been shown to be beneficial to apply self-attention multiple times with different query-key-value triplets, which is called ``Multi-Head Attention'' \citep{vaswani_attention_2017}.

\subsubsection{Self-supervised learning of images}

Self-supervised learning describes a methodology where no external labels are required to train a model, but the training objective is generated from the data itself in the form of a proxy task. In general, this enables training based on large unlabelled data sets.
\citet{chen_empirical_2021} present an overview study of self-supervised vision transformers.

In this work, we use one specific self-supervised learning approach called ``DINO'' (self \textbf{di}stillation with \textbf{no} labels) \citet{caron_emerging_2021}.
It adapts the concept of knowledge distillation, where a student model is learning from a teacher model. Both models share the same architecture, while the teacher is dynamically created from the exponential moving average of the student's parameters. 
During a training step, the two models receive different augmentations of the same input image, while the learning objective is to minimize the divergence of the two outputted probability distributions.
As a result, the student model learns the relevant information in an image in a self-supervised way.

The DINO framework works particularly well with ViTs but can, in principle, also be applied to other model architectures, such as CNNs. 
The authors have shown that DINO ViTs explicitly capture the scene layout in the form of object boundaries, which does neither apply to CNNs nor supervised ViTs. Further experiments have shown that their model can achieve competitive classification results on benchmark data sets when combined with simple linear classification models. 
These properties let us hypothesize that ViTs trained with the DINO method are particularly suitable as feature extractors. While supervised methods might be prone to being hyper-optimized on the task they are trained on, we will evaluate if DINO features are superior in capturing the general task-independent information of an image.
In this study, we use pre-trained DINO ViTs as ``off-the-shelf'' feature generators, which means we use them on domain data sets without retraining or fine-tuning.

\subsection{Experimental setup}

The classification experiments described in this study were implemented using PyTorch \citep{paszke_pytorch_2019} version 1.9.1 and scikit-learn \citep{pedregosa_scikit-learn_2011} version 0.24.2.
To address the robustness of the reported results, all training runs were executed at least five times with different random initialization.
For all experiments, we use the same training (64\%), validation (16\%), and test (20\%) splits.
To evaluate how different models perform on different data set sizes, we artificially reduced the available training data for some runs by sampling class-balanced subsets from the original data set using weighted random sampling. The validation and the test split stay at their original size in those cases.

\subsubsection{Baseline: Convolutional Neural Networks}

As a baseline, we evaluate several state-of-research CNNs; all pre-trained on the ImageNet data set \citep{deng_imagenet_2009}. The models included in this study are VGG-11 \citep{simonyan_very_2014}, AlexNet \citep{krizhevsky_one_2014}, DenseNet \citep{huang_densely_2017}, SqueezeNet \citep{iandola_squeezenet_2016}, and four variations of ResNet \citep{he_deep_2016} with different numbers of residual blocks.

We use pre-trained weights for the feature extractor block in the CNN, and the classifier is initialized with random weights. Two different types of transfer learning with CNNs were examined: a) train the whole model on the domain data set, and b) only train the classifier on the domain data set (see \Cref{fig:concept} for reference).

We trained with a batch size of 100 samples using Stochastic Gradient Descent as an optimization method. Further, we follow well-established practices when training neural networks: 
The initial learning rate and momentum are 0.0001 and 0.9, respectively.
The learning rate is decreased once the model converges towards its best state: we reduce it by a factor of 10 every time the validation loss does not decrease for 15 subsequent epochs.
To avoid overfitting on the training data, training is stopped once the validation loss does not decrease for 30 epochs in a row, while the final model weights are loaded from the checkpoint where convergence started.

\subsubsection{Dino ViTs + shallow classifiers}

\begin{figure*}[ptb!]
    \centering
    \includegraphics[width=\textwidth]{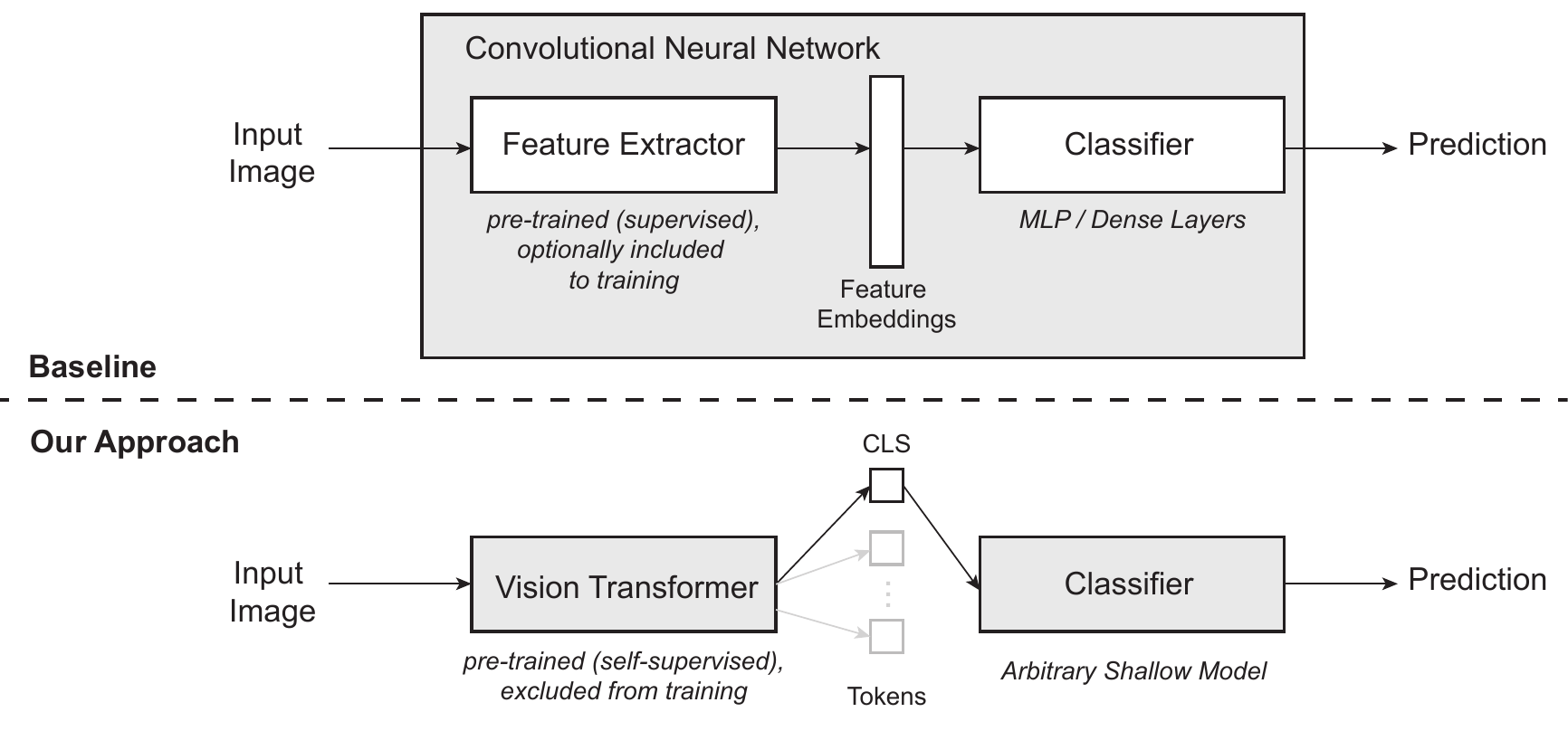}
    \caption{High-level comparison between the standard approach for machine learning--based image classification and the method proposed in this work. We use a pre-trained Vision Transformer for the task of feature extraction. The generated CLS token can be seen as the equivalent of CNN embeddings. Since the training of the ViT and the classifier are decoupled, we can use any shallow model for class prediction.}
    \label{fig:concept}
\end{figure*}
\vfill
\Cref{fig:concept} (bottom) shows a high-level overview of our approach compared to the baseline procedure.
We use DINO ViTs that are pre-trained on ImageNet as provided by the original authors\footnote{\url{https://github.com/facebookresearch/dino}} to generate the CLS token from an image. Different ViT configurations are tested: We test the small (``ViT-S'', ~21 million parameters, 384 token dimensions) and the base architecture (``ViT-B'', ~85 million parameters, 768 token dimensions). Both are combined either with an 8x8 or 16x16 pixel patch size.

Based on the generated CLS token, we fit a set of the most popular shallow machine learning models for classification tasks.
The models included in this study are K-nearest Neighbors (KNN), Logistic Regression (LR), Support Vector Machine (SVM), Random Forest (RF), XGBoost \citep{chen_xgboost_2016}, and Multilayer Perceptron (MLP), which is a small neural network.
All of those are classical and well-established techniques in the machine learning community and have been shown to work well on tabular data.

\subsection{Visualizing Feature Embeddings}

Even though feature embeddings generated by deep learning models have significantly fewer dimensions than the original images, their dimensions are still in the hundreds. To visualize the distribution of samples, we use \textit{unsupervised} machine learning methods for dimensionality reduction. Those methods aim to decrease the number of dimensions while maintaining the most important information. In this study, we used Principal Component Analysis (PCA) as a linear method and UMAP \citep{mcinnes_umap_2018} from the family of non-linear methods.
\section{Results and Discussion}

\subsection{Classification results on the full data sets}

We evaluate the predictive performance of CNNs and our proposed method in classifying the banana fruit images.
\Cref{tab:results_fayoum_cnn} comprises the test set classification accuracies for banana ripeness classification using CNNs. The classification works similarly well, independent of whether the entire model or only the classification layers are trained. This is plausible since the task is comparably simple and mainly color-based. The best classification accuracy for CNNS (0.941) is the same as for the best ViT-based approach (see \Cref{tab:results_fayoum_dino}).
We hypothesize that even higher accuracies for this data set for both CNN and DINO ViT are prevented by the annotation blur and possibly wrongly labeled data points we described in \Cref{sec:datasets}. It can be assumed that all evaluated approaches can solve the task almost perfectly if labels are accurate. However, having a certain degree of noise and inaccuracy in labels comes with most real-world data sets.

\begin{table*}[t!]
\caption{Test set classification accuracies for both data sets included in this study. We show the results for CNNs ((a) and (b)) and ViTs combined with shallow classifiers ((c) and (d)). All deep models are pretrained on ImageNet.}
\label{tab:results_fayoum}
  \begin{subtable}[b]{.48\textwidth}\centering
\caption{CNN results for the Fayoum banana data set. We report both accuracies when all weights are trained and those when only the dense classification layers are trained. The \fr{first} and \sr{second} best scores for each variation are highlighted.}\label{tab:results_fayoum_cnn}
{\begin{tabularx}{\columnwidth}{LCC}
\toprule
Model &     Full model & Classifier only \\
\midrule
AlexNet    &  \sr{0.930 ± 0.015} &   \fr{0.941 ± 0.015} \\
DenseNet   &  0.926 ± 0.029 &   0.907 ± 0.013 \\
ResNet-101 &  0.922 ± 0.008 &   0.893 ± 0.015 \\
ResNet-152 &  0.922 ± 0.024 &   \sr{0.941 ± 0.020} \\
ResNet-18  &  0.926 ± 0.013 &   0.904 ± 0.024 \\
ResNet-50  &  \fr{0.933 ± 0.021} &   0.907 ± 0.000 \\
SqueezeNet &  0.915 ± 0.021 &   0.933 ± 0.010 \\
VGG        &  0.904 ± 0.027 &   0.885 ± 0.020 \\
\bottomrule
\end{tabularx}}
\end{subtable}
  \hfill
\begin{subtable}[b]{.48\textwidth}\centering
\caption{CNN results for the CASC IFW apple data set. We report both accuracies when all weights are trained and those when only the dense classification layers are trained. The \fr{first} and \sr{second} best scores for each variation are highlighted.}\label{tab:results_cascifw_cnn}
{\begin{tabularx}{\columnwidth}{LCC}
\toprule
Model &     Full model & Classifier only \\
\midrule
AlexNet    &  0.938 ± 0.004 &   0.866 ± 0.001 \\
DenseNet   &  0.949 ± 0.003 &   0.847 ± 0.004 \\
ResNet-101 &  0.950 ± 0.003 &   0.850 ± 0.002 \\
ResNet-152 &  0.944 ± 0.004 &   0.859 ± 0.001 \\
ResNet-18  &  0.944 ± 0.003 &   0.842 ± 0.002 \\
ResNet-50  &  0.949 ± 0.003 &   0.846 ± 0.005 \\
SqueezeNet & \sr{0.952 ± 0.002} &   \fr{0.879 ± 0.004} \\
VGG        &  \fr{0.958 ± 0.002} &   \sr{0.872 ± 0.005} \\
\bottomrule
\end{tabularx}}
\end{subtable}

  \medskip

  \begin{subtable}[b]{.48\textwidth}\centering
\caption{Results for the Fayoum banana data set of shallow classifiers based on embeddings of two different DINO ViTs (base and small model). The transformers are never trained on the domain data set. The \fr{first} and \sr{second} best score are highlighted.}
\label{tab:results_fayoum_dino}
\begin{tabularx}{\columnwidth}{LCC}
\toprule
\small Transformer \textrightarrow &     \small DINO ViT-B/8 &     \small DINO ViT-S/8 \\
\small Classifier \textdownarrow & & \\
\midrule
KNN   &  0.922 ± 0.008 &  0.904 ± 0.015 \\
LR     &  0.930 ± 0.008 &  0.900 ± 0.010 \\
MLP          &  0.933 ± 0.010 &  0.915 ± 0.010 \\
RF  &  \sr{0.937 ± 0.010} &  0.915 ± 0.021 \\
SVM                    &  0.919 ± 0.017 &  0.904 ± 0.008  \\
XGBoost          &  0.893 ± 0.024 &  \fr{0.941 ± 0.030} \\
\bottomrule
\end{tabularx}
\end{subtable}
  \hfill
  \begin{subtable}[b]{.48\textwidth}\centering
\caption{Results for the CASC IFW apple data set of shallow classifiers based on embeddings of two different DINO ViTs (base and small model). The transformers never trained on the domain data set. The \fr{first} and \sr{second} best score are highlighted.}\label{tab:results_cascifw_dino}
{\begin{tabularx}{\columnwidth}{LCC}
\toprule
\small Transformer \textrightarrow &     \small DINO ViT-B/8 &     \small DINO ViT-S/8 \\
\small Classifier \textdownarrow & & \\
\midrule
KNN   &  0.873 ± 0.002 &  0.903 ± 0.004 \\
LR &  0.917 ± 0.002 &  0.917 ± 0.003 \\
MLP &  0.934 ± 0.002 &  0.930 ± 0.004 \\
RF  &  0.872 ± 0.005 &  0.879 ± 0.002 \\
SVM &  \fr{0.950 ± 0.000} &  \sr{0.936 ± 0.001} \\
XGBoost &  0.925 ± 0.003 &  0.910 ± 0.004 \\
\bottomrule
\end{tabularx}}
\end{subtable}
\end{table*}

\Cref{tab:results_cascifw_cnn} shows the CNN results for the apple fruit data set.
The classification accuracy is significantly higher when training the full model compared to when only training the classification layers. This increased accuracy is likely the case because we are looking for specific and local features in the image which are not captured by the pre-trained model.

As shown in \Cref{tab:results_cascifw_dino}, the best ViT-based model (a Support Vector Classifier based on DINO ViT-B/8 embeddings) performs less than one percent worse than the best CNN (0.950 vs. 0.958). At the same time, it significantly outperforms CNNs in terms of accuracy, where only the classification layers are trained. These are remarkable results since we did not train the ViT on the domain data set. It shows that the pre-trained DINO ViT can capture domain-specific and local features without having seen comparable images before. Consequently, for advanced classification tasks, DINO ViTs could either achieve similar results with less training effort (compared to fully trained CNNs) or better results with comparable training effort (compared to CNNs where only the classifier is trained).

These results evoke the impression that DINO ViT embeddings are generally more accurate than supervised CNN embeddings in capturing the essential features of images when models are pre-trained on general-purpose data sets. To confirm this claim, we ran another set of experiments where we combined the best models from previous results with various pre-trained ViTs and CNNs, including a DINO-trained ResNet-50 for comparison. The results are shown in \Cref{tab:encoders}. Again, the best accuracies are achieved by DINO ViT-based configurations. This implies that the improved accuracy stems from the feature quality rather than from a possibly superior classifier.

It is worth mentioning that previous machine learning research using the same two data sets achieves higher overall accuracies \citep{ismail_real-time_2022}. The main difference in their approach is sophisticated and highly optimized image pre-processing. We are not benchmarking our results with theirs, as this research focuses on implementation simplicity rather than maxing out predictive performance for the given data sets.

\begin{table}[tb!]
\centering
\caption{Test set classification accuracies of the best shallow classifier from previous experiments in combination with feature embeddings from both ViTs and CNNs. The \fr{first} and \sr{second} best score for each data set are highlighted.}
\begin{tabular}{lcc}
\toprule
\textit{Data set} &  \textit{Fayoum Banana} &       \textit{CASC IFW} \\
Classifier \textrightarrow &       XGBoost &       SVM \\
Embeddings \textdownarrow & &  \\
\midrule
DINO ViT-B/8    &  0.893 ± 0.024 &  \fr{0.950 ± 0.000} \\
DINO ViT-S/8    &  \fr{0.941 ± 0.030} &  0.936 ± 0.001 \\
DINO ViT-B/16   &  0.919 ± 0.010 &  \sr{0.947 ± 0.002} \\
DINO ViT-S/16   &  0.896 ± 0.025 &  0.909 ± 0.002 \\
DINO ResNet-50  &  0.889 ± 0.029 &  0.893 ± 0.005 \\
ResNet-18       &  0.870 ± 0.029 &  0.863 ± 0.003 \\
ResNet-50       &  0.911 ± 0.008 &  0.869 ± 0.003 \\
ResNet-101      &  \sr{0.930 ± 0.030} &  0.889 ± 0.003 \\
ResNet-152      &  0.926 ± 0.029 &  0.882 ± 0.004 \\
VGG             &  0.874 ± 0.024 &  0.836 ± 0.004 \\
\bottomrule
\end{tabular}
\label{tab:encoders}
\end{table}

\subsection{Classification results with reduced training data}

\begin{figure*}[tb!]
    \centering
    \includegraphics[width=\textwidth]{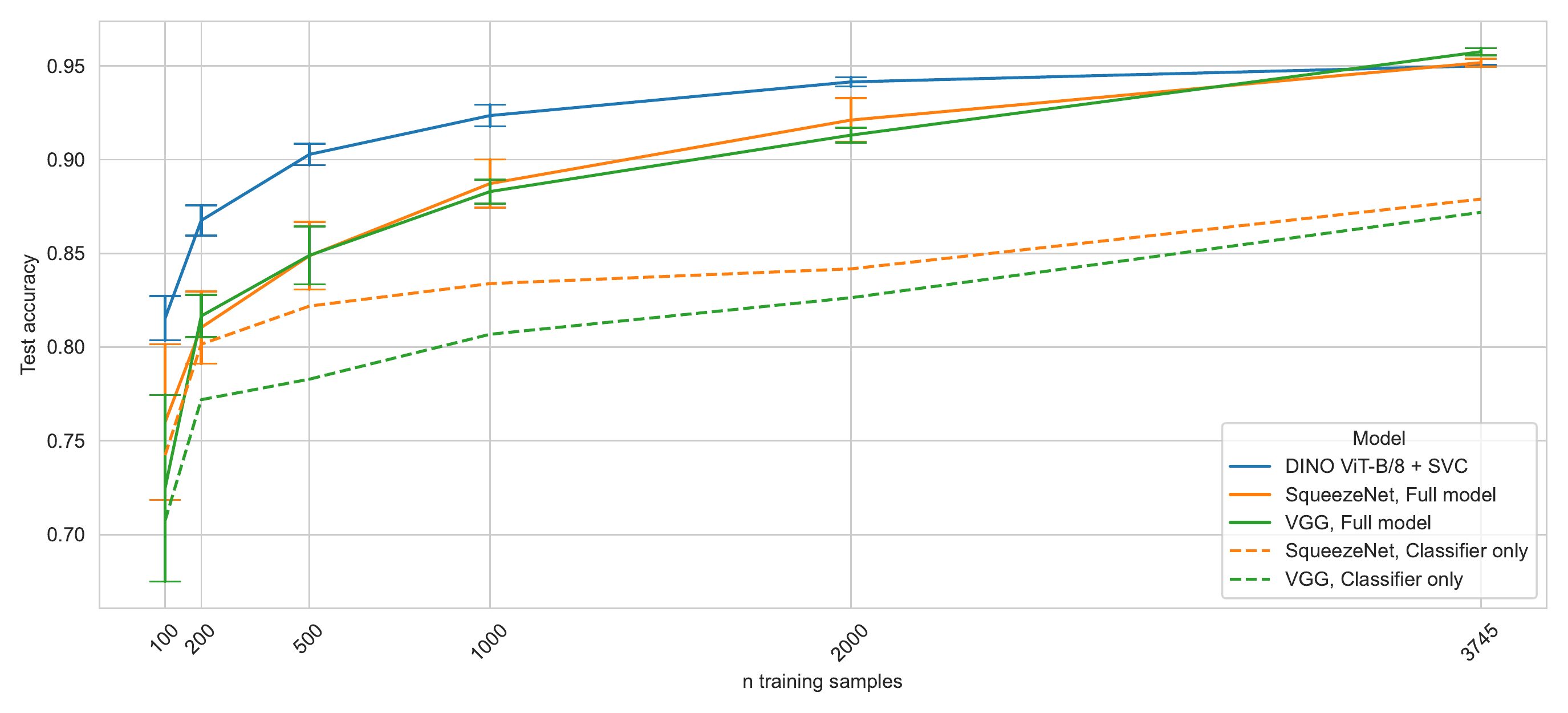}
    \caption{Test accuracies on the CASC~IFW data set for different numbers of available training samples. Error bars represent the standard deviation of experimental results w.r.t. random initialization. The ViT approach performs significantly better when training data is scarce. $90\%$ accuracy is reached with three times less training data compared to the best-performing CNNs.}
    \label{fig:acc-samples}
\end{figure*}

The fact that we need to train any deep neural network on our domain data sets to achieve good results naturally leads to the hypothesis that our approach works better on small data sets than CNNs. To test this hypothesis, we chose the best-performing models from previous experiments to examine their performance when training data is scarce. 
We trained the models on randomly sampled subsets of the training split while the validation and test set stayed constant and were the same as in previous runs. \Cref{fig:acc-samples} shows the models' performances as a function of the available training data using the CASC~IFW apple data set. We can observe that DINO ViT-based classifiers perform significantly better on small data sets. A test accuracy higher than 0.9 can be achieved with only 500 samples, while the two CNNs need at least 1500 samples to reach that benchmark.
The same experiment for the Fayoum Banana data set can be found in \ref{sec:appendixB}. However, the results are harder to interpret since downsampling an already very small data set leads to sample sizes where all models perform equally poorly.

\subsection{Embedding visualization}
\label{sec:embedding_vis}

\begin{figure*}[tb!]
    \centering
    \includegraphics[width=.9\textwidth]{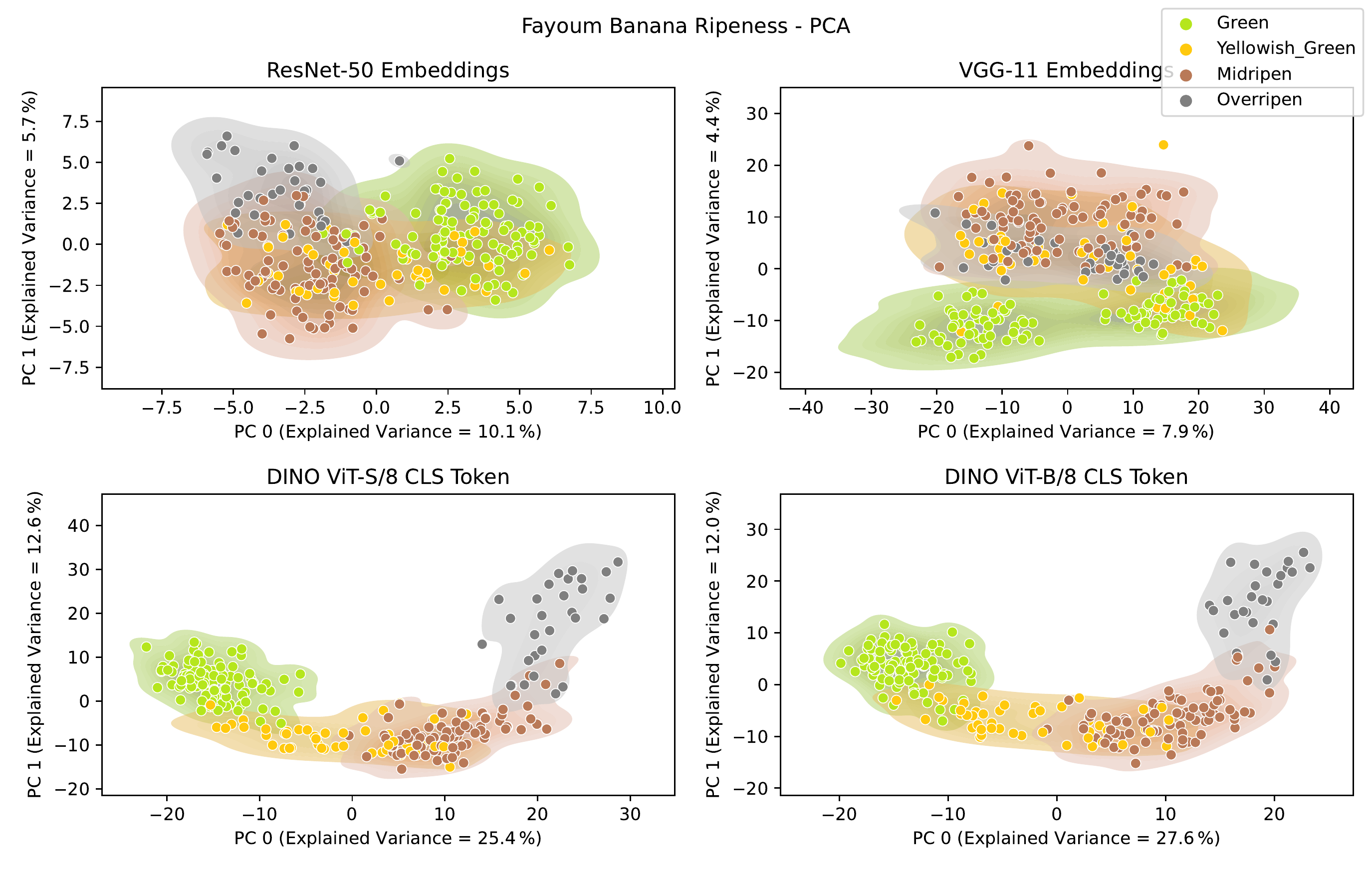}
    \caption{Latent representation of the Fayoum banana ripeness data set in two dimensions (reduced by PCA) for ResNet50 (top-left), VGG-11 (top-right), DINO ViT-S/8 (bottom-left), and DINO ViT-B/8 (bottom-right). All models are pre-trained on ImageNet, and there was no training on the domain data set. The ViT models seem superior in capturing the ordinal property of the four classes. Overlaps between the classes can be traced back to imprecise labeling.}
    \label{fig:pca-fayoum}
\end{figure*}

This section examines the differences between CNN and DINO ViT embeddings by visualizing them in one or two dimensions using dimensionality reduction methods. Subsequently, we only show the results based on PCA. The reader is encouraged to explore visualizations of further principal components and visualizations based on the UMAP algorithm in \ref{sec:appendixC}. 
It is important to note that dimensionality reduction is done unsupervised, which means that the information on class labels is added a posteriori to the visualizations. As representatives for CNNs, we chose a pre-trained VGG network since it is one of the strongest models for our data sets from previous experiments and ResNet-50 since it is a popular and widely used model. These models are compared to a ViT-S/8 and a ViT-B/8 model. \Cref{fig:pca-fayoum} shows the distribution of banana samples in two dimensions. Although low-dimensional CNN embeddings cluster the classes to a certain degree, there is a substantial overlap between them. A closer look at VGG embeddings shows that the model captures the orientation of bananas in the first PCA dimension and divides them into two classes (negative and positive x). On the other hand, DINO ViT embeddings show a clearer separation of clusters and, at the same time, capture the ordinal character of ripeness classes. The largest overlap exists between the ``yellowish green'' and the ``midripen'' class, which can be traced back to the aforementioned imprecise labeling.
One reason for this clear class separation is that there are not many signals in the image apart from the fruit's color and shape since lighting conditions and background are constant.

\begin{figure*}[tb!]
    \centering
    \includegraphics[width=.9\textwidth]{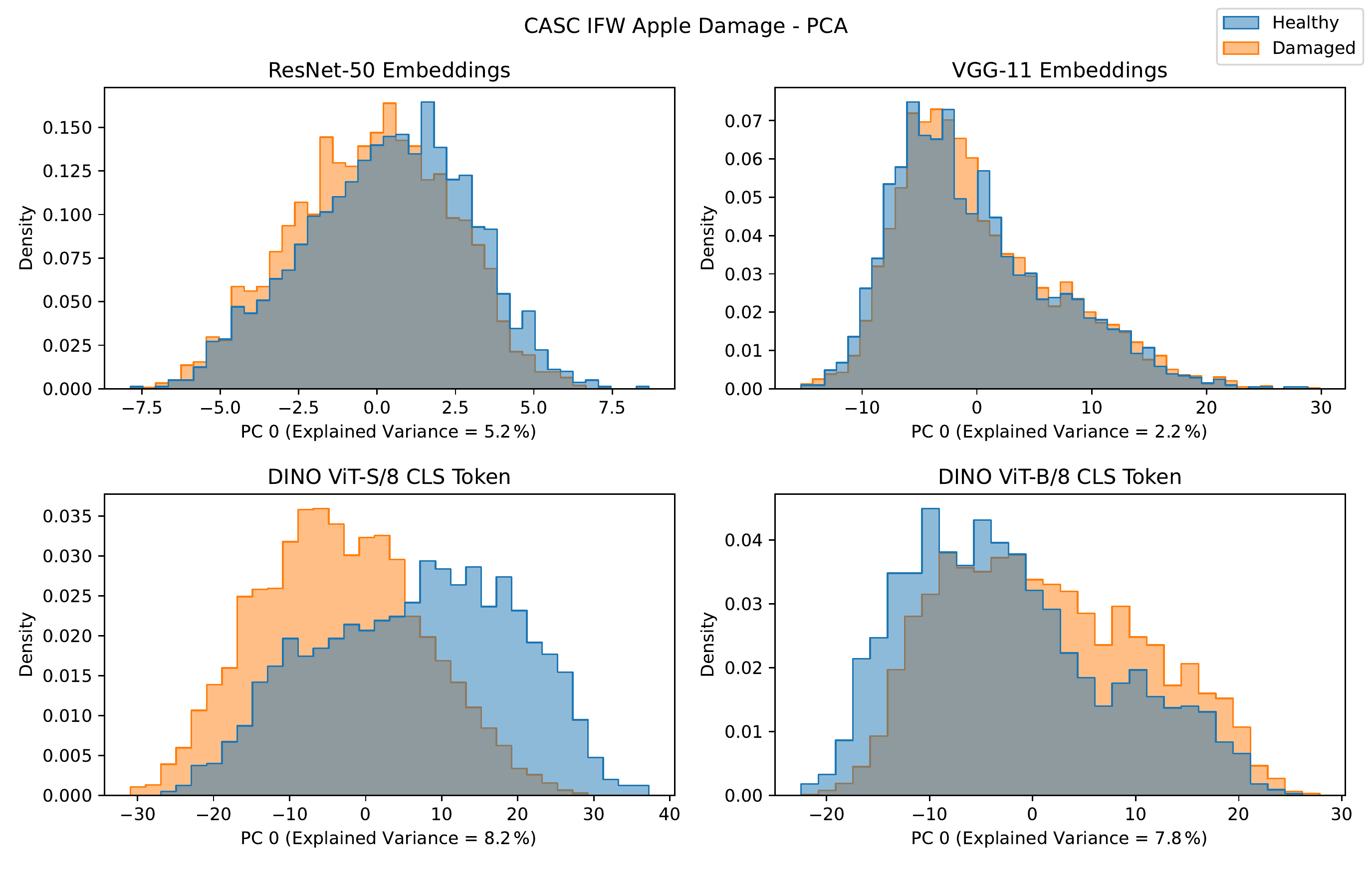}
    \caption{Latent representation of the CASC IFW apple defect data set in one dimension (reduced by PCA) for ResNet50 (top-left), VGG-11 (top-right), DINO ViT-S/8 (bottom-left), and DINO ViT-B/8 (bottom-right). All models are pre-trained on ImageNet, and there was no training on the domain data set. Even though defects are a non-dominant visual feature, there is a notable shift in the distribution of the first PCA component of ViT embeddings, especially for DINO ViT-8.}
    \label{fig:pca-cascifw}
\end{figure*}

For reasons of clarity, we only show the distribution of the first principal component of apple image embeddings in \Cref{fig:pca-cascifw}. The first principal component of CNN embeddings generates similar distributions for healthy and damaged apples. We hypothesize that more dominant features, such as the fruit's color, changes in spectra, or background signals, are more salient in low dimensions. Further analysis of the second and third components (see \ref{sec:appendixC}) indicates that local defects are captured there. However, the separation of the classes is rather marginal. 
In comparison, one can observe a clear distribution shift of the healthy/damaged classes using the first PC of DINO ViT embeddings.
This observation allows two possible interpretations: First, DINO ViT embeddings seem more robust against noise, such as lighting and background. Second, these embeddings lead to a higher degree of linearization in the embedding space.
In any case, these visualizations support the findings of the classification experiments: DINO ViT features are a superior low-dimensional representation of images compared to CNNs when pre-trained models are being used as off-the-shelf feature generators.

\section{Conclusion and Outlook}

We presented a simplified development procedure for image-based machine learning for visual fruit quality assessment. It is particularly suitable for domains with low availability of both data and computational resources.
%
%
%
%

Our main contribution is the facilitation of the development process of the ML model compared to current state-of-research approaches. 
We show how effective machine learning models can be established with minimal development effort on a standard computer without GPU access. 
In addition, our approach maintains a high classification accuracy with small training data sets. 
This can democratize the use of machine learning in decentralized food supply chains. Potential users are cold store operators or companies serving farmer-producer organizations, who often rely on smartphone images for fruit quality assessment. 
%
This research contributes toward ``Green AI'' which aims to make AI systems more efficient from a computational standpoint and more inclusive for researchers and engineers worldwide by not being dependent on the availability of enhanced computational resources \citep{schwartz_green_2020}.

Future research could focus on validating our findings on more difficult classification tasks. We do not address whether and how far fine-tuning pre-trained Vision Transformers could outperform current state-of-research methods in achieving the highest possible predictive performance.
Many food quality applications do not rely on image labels but require the localization and/or quantification of defects. Future studies could address this by adapting the presented approach to object detection or image segmentation tasks.
Our visualizations of feature embeddings using PCA underline the high expressiveness of DINO ViT embeddings. This is likely a result of the relevant features of images being largely linearized in the embedding space. This property indicates that DINO ViT embeddings may be particularly suitable for integration with other data sources. One could combine these embeddings (or some principal components) with other data types in a common tabular form. This could be, for instance, sensor or meta data from food supply chains to predict the end quality at arrival.


\section*{CRediT authorship contribution statement}
\textbf{Manuel Knott:} Conceptualization, Methodology, Software, Formal analysis, Investigation, Writing -- Original Draft, Visualization.
\textbf{Fernando Perez-Cruz:} Conceptualization, Methodology, Resources, Writing -- Review \& Editing, Supervision, Funding acquisition.
\textbf{Thijs Defraeye:} Conceptualization, Resources, Writing -- Review \& Editing, Supervision, Project administration, Funding acquisition.

\section*{Declaration of competing interests}
The authors declare that they have no known competing financial interests or personal relationships that could have influenced the work reported in this paper.

\section*{Acknowledgments}
This work was funded partially by the data.org Inclusive Growth and Recovery Challenge grant ``Your Virtual Cold Chain Assistant'', supported by The Rockefeller Foundation and the Mastercard Center for Inclusive Growth. The funder was not involved in the study design, collection, analysis, interpretation of data, the writing of this article or the decision to submit it for publication. This manuscript has been released as a pre-print at arXiv. We acknowledge the support of Daniel Onwude for critical commenting on the work.

\section*{Supplementary materials}

The source code of all experiments can be accessed on Github\footnote{\url{https://github.com/manuelknott/DINO-ViT_fruit_quality_assessment}}.

\bibliographystyle{elsarticle-harv}
\bibliography{references} 

\clearpage
\appendix
\onecolumn

\section{Example Images}
\label{sec:appendixA}
\setcounter{figure}{0}
\setcounter{table}{0}

\begin{figure}[ht!]
\centering
\begin{subfigure}[b]{\textwidth}
   \includegraphics[width=.95\linewidth]{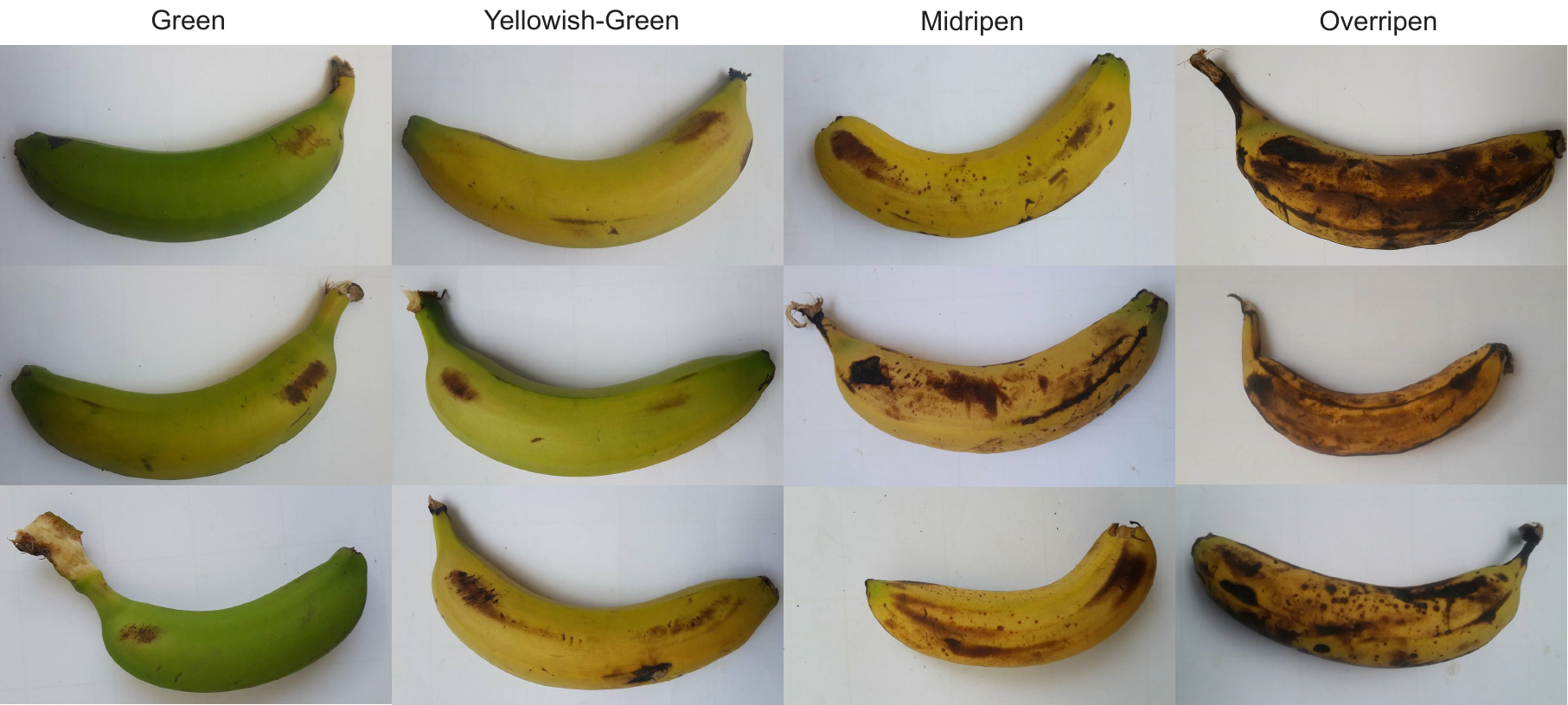}
   \caption{Example of the Fayoum banana ripeness data set \citep{mazen_ripeness_2019}, where three images of each class are depicted.}
   \label{fig:fayoum_examples} 
\end{subfigure}

\begin{subfigure}[b]{\textwidth}
   \includegraphics[width=.95\linewidth]{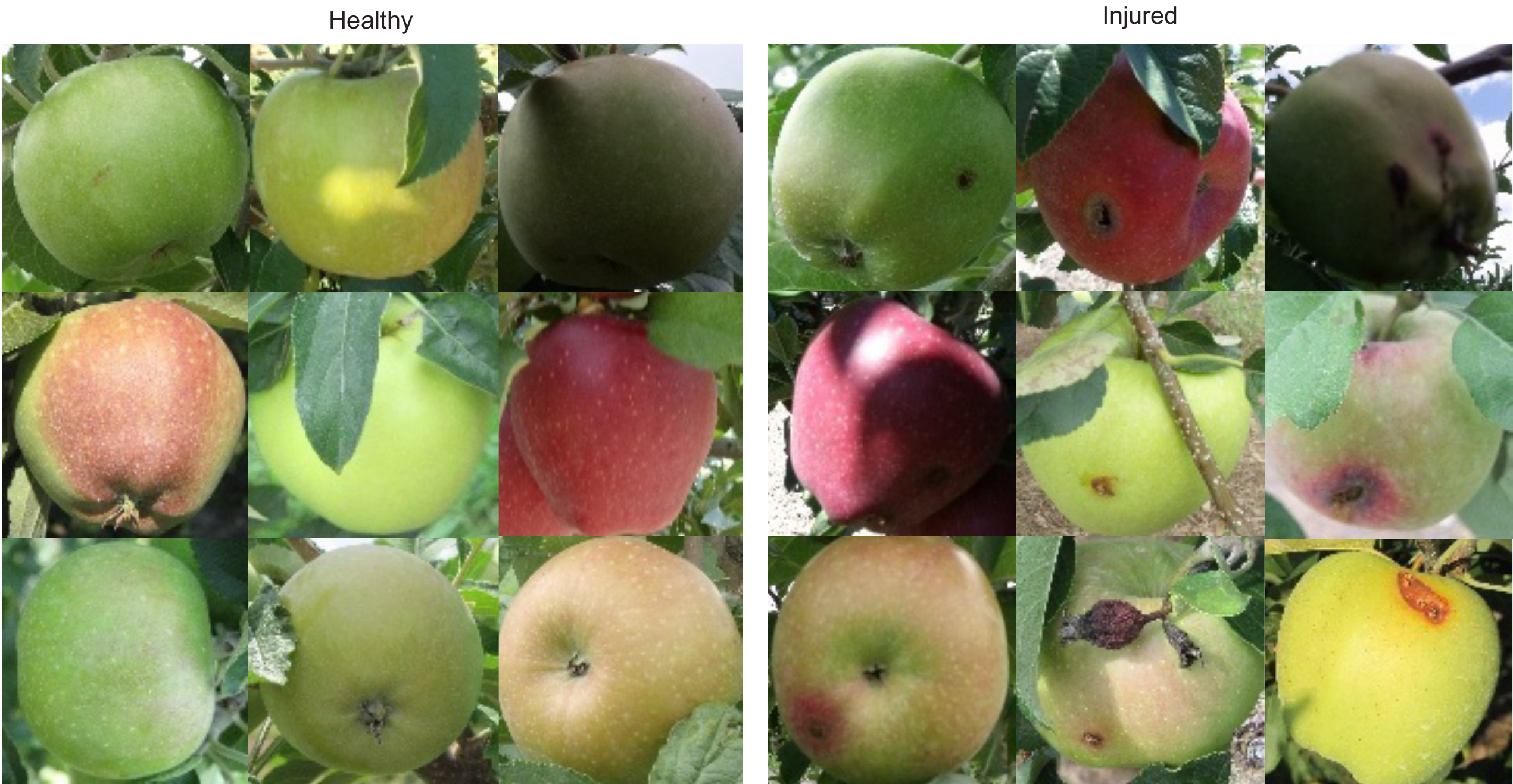}
   \caption{Example of the CASC IFW apple damage data set, where nine images per class are depicted.}
   \label{fig:cascifw_examples}
\end{subfigure}

\caption{Example images from the two data sets used in this research.}
\end{figure}

\pagebreak
\section{Sample size experiments for Fayoum data set}
\label{sec:appendixB}
\setcounter{figure}{0}
\setcounter{table}{0}

\begin{figure*}[ht!]
    \centering
    \includegraphics[width=\textwidth]{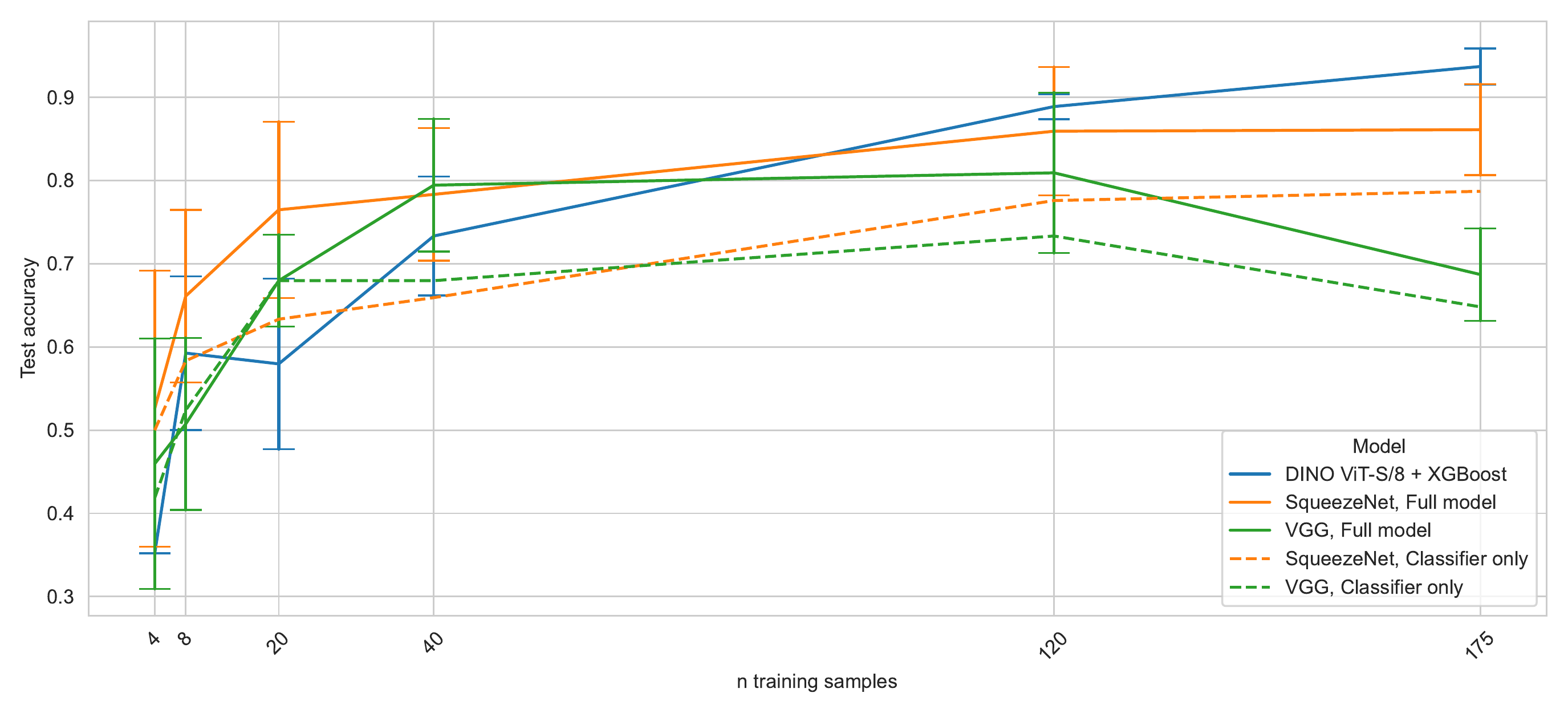}
    \caption{Test accuracies on the Fayoum Banana data set for different numbers of available training samples. Error bars represent the standard deviation of experimental results w.r.t. random initialization. In Contrast to the results shown in \Cref{fig:acc-samples}, the ViT-based approach is not superior with a low number of samples. }
    \label{fig:acc-samples-fayoum}
\end{figure*}

\section{Additional Embedding Visualizations}
\label{sec:appendixC}
\setcounter{figure}{0}
\setcounter{table}{0}

\begin{figure*}[ht!]
    \centering
    \includegraphics[width=.75\textwidth]{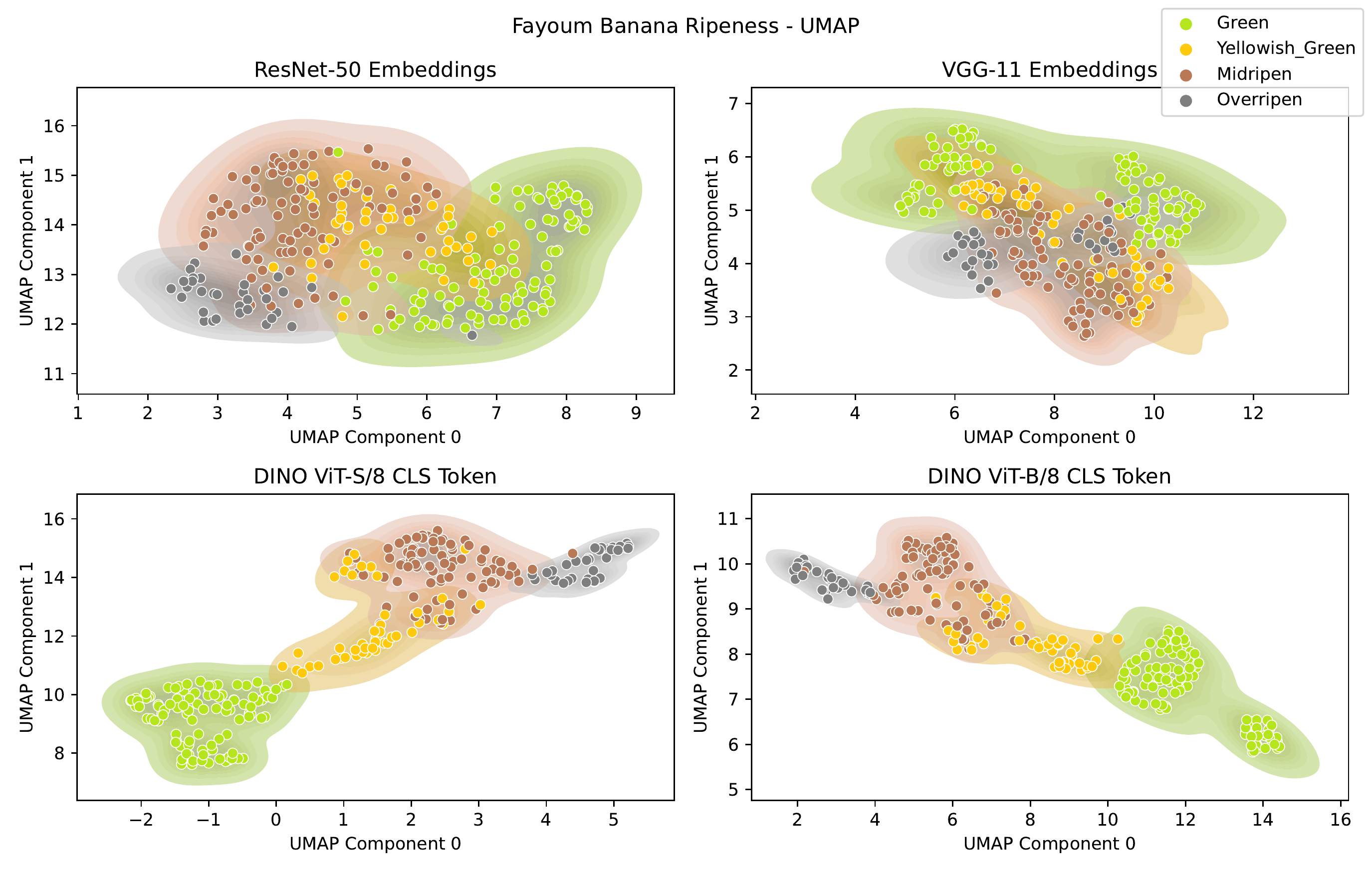}
    \caption{Latent representation of the Fayoum banana ripeness data set in two dimensions (reduced by UMAP) for Resnet50 (top-left), VGG (top-right), DINO ViT-S/8 (bottom-left), and DINO ViT-B/8 (bottom-right). All models are pre-trained on ImageNet and there was no training on the domain data set.}
\end{figure*}


\begin{figure*}[ht!]
    \centering
    \includegraphics[width=.75\textwidth]{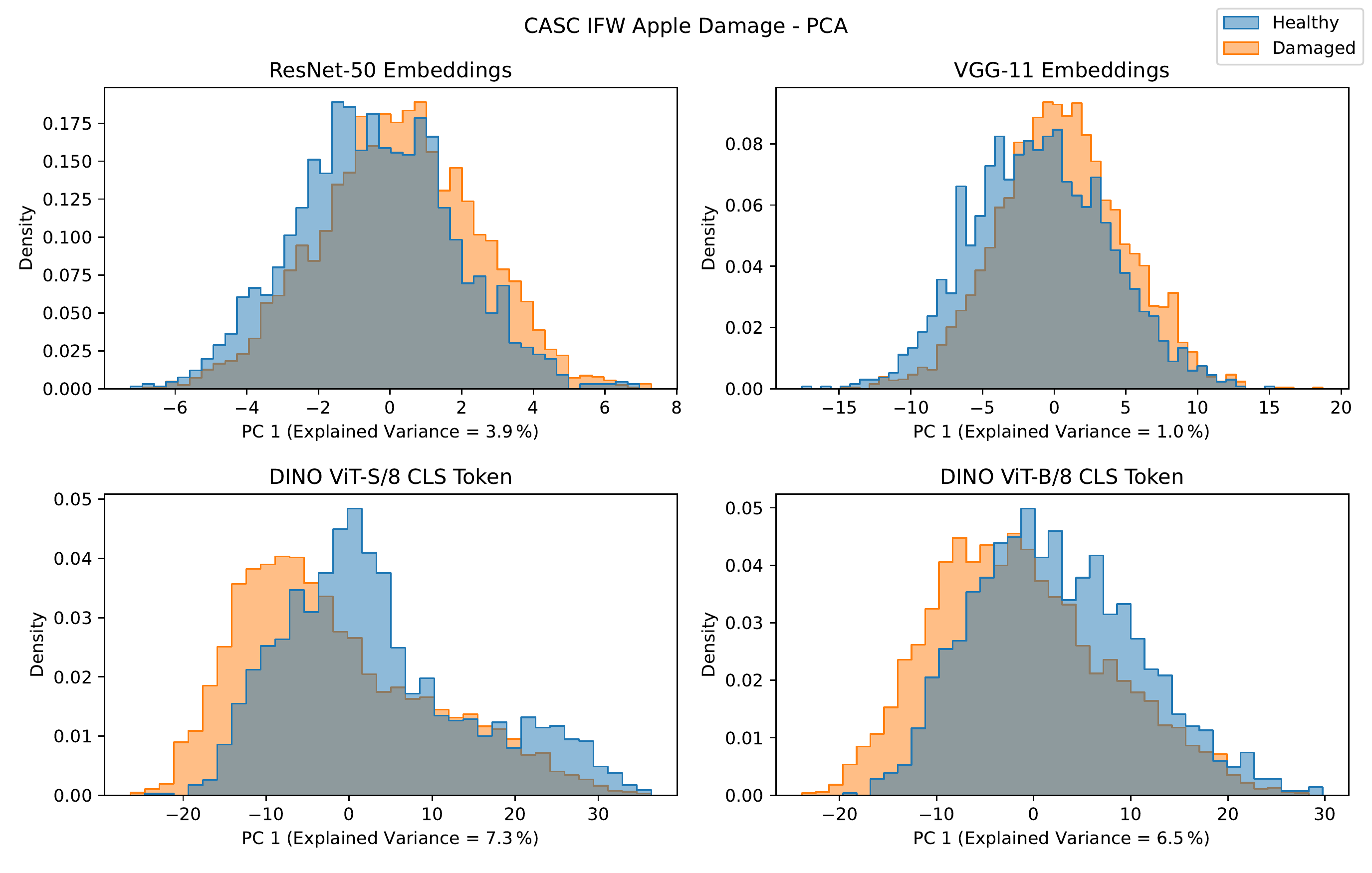}
    \caption{Second Principal Component (PC1) of the latent representations generated by different models for the CASC IFW apple defect data set. See \Cref{fig:pca-cascifw} for further explanation.}
\end{figure*}

\begin{figure*}[ht!]
    \centering
    \includegraphics[width=.75\textwidth]{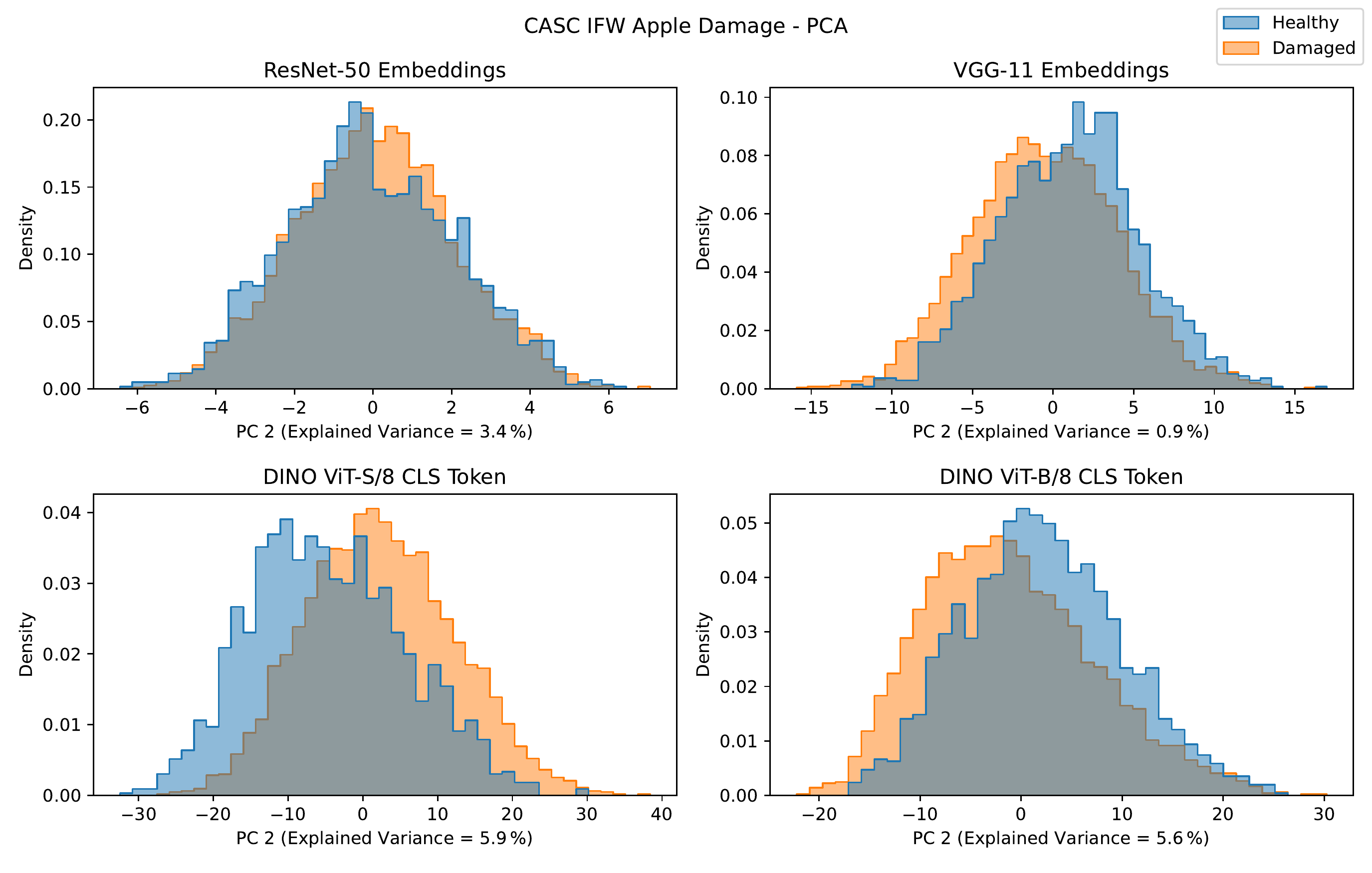}
    \caption{Third Principal Component (PC2) of the latent representations generated by different models for the CASC IFW apple defect data set. See \Cref{fig:pca-cascifw} for further explanation.}
\end{figure*}

\begin{figure*}[ht!]
    \centering
    \includegraphics[width=.75\textwidth]{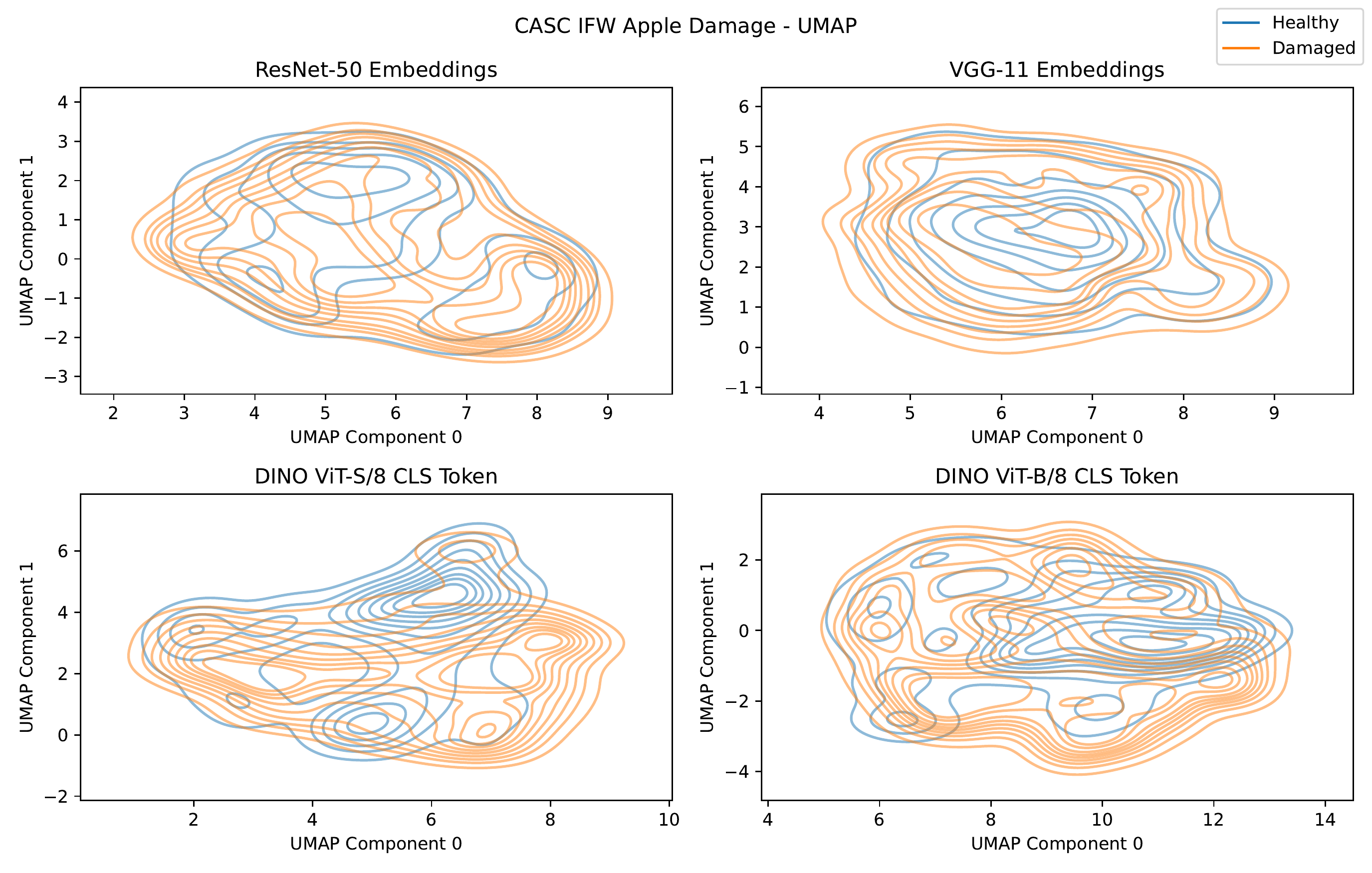}
    \caption{Latent representation of the CASC IFW apple defect data set in two dimensions (reduced by UMAP) for Resnet50 (top-left), VGG (top-right), DINO ViT-S/8 (bottom-left), and DINO ViT-B/8 (bottom-right). Data points are omitted for clarity. All models are pre-trained on ImageNet and there was no training on the domain data set.}
\end{figure*}

\renewcommand{\thefigure}{\arabic{figure}}
\renewcommand{\thetable}{\arabic{table}}


\end{document}